\documentclass[pdflatex,sn-mathphys]{sn-jnl}

\jyear{2023}%

\theoremstyle{thmstyleone}%
\theoremstyle{thmstyletwo}%

\theoremstyle{thmstylethree}%

\usepackage{adjustbox}
\begin{document}

\title[Efficient Training of One Class Classification
- SVMs]{Efficient Training of One Class Classification - SVMs}


\author*[1]{\fnm{Isaac A.} \sur{Yowetu}}\email{isaacamornorteyyowetu@gmail.com}

\equalcont{These authors contributed equally to this work.}

\author[2]{\fnm{Nana K.} \sur{Frempong}}\email{nkfrempong@gmail.com}
\equalcont{These authors contributed equally to this work.}

\affil[1]{\orgdiv{Department of Mathematics}, \orgname{Kwame Nkrumah University of Science and Technology}, \orgaddress{\street{Kumasi}, \city{Ashanti Region}, \postcode{00233}, \country{Ghana}}}

\affil[2]{\orgdiv{Department of Statistics and Actuarial Science}, \orgname{Kwame Nkrumah University of Science and Technology}, \orgaddress{\street{Kumasi}, \city{Ashanti Region}, \postcode{00233}, \country{Ghana}}}



\abstract{This study examines the use of a highly effective training method to conduct one-class classification. The existence of both positive and negative examples in the training data is necessary to develop an effective classifier in common binary classification scenarios. Unfortunately, this criteria is not met in many domains. Here, there is just one class of examples. Classification algorithms that learn from solely positive input have been created to deal with this setting. In this paper, an effective algorithm for dual soft-margin one-class SVM training is presented. Our approach makes use of the Augmented Lagrangian (AL-FPGM), a variant of the Fast Projected Gradient Method. The FPGM requires only first derivatives, which for the dual soft margin OCC-SVM means computing mainly a matrix-vector product. Therefore, AL-FPGM, being computationally inexpensive, may complement existing quadratic programming solvers for training large SVMs. We extensively validate our approach over real-world datasets and demonstrate that our strategy obtains statistically significant results.}

\maketitle
\section{Introduction}\label{sec1}
Learning algorithms consider the availability of both positive and negative examples in common binary classification tasks. Sometimes a strong requirement like this is needed but it does not work in the context of an application for real-world application. In actuality, labeling data is an expensive, time-consuming task that necessitates a high level of domain knowledge. In some cases, this operation is quick, but usually, defining reliable labels for each data example is a hard task \citep{ienco2016positive}.

The goal of one-class classification (OCC) methods is to create classification models when the negative class is either nonexistent, poorly sampled, or poorly specified \citep{Khan2014OneclassCT}. Thus, this technique creates  class boundaries only with the knowledge of positive class. Examples of one-class classification include outlier detection and novelty detection, where the outlier elements are identified independently of all the other data elements. This One-Class classification problem happens in a variety of situations, including:
\begin{itemize}
    \item \textbf{Outlier Detection: }The objective is to find samples from an unlabeled dataset that are outliers. Outliers in the training set are observations that deviate significantly from the others. Outlier estimators ignore the deviating observations and attempt to fit the majority of the training data under the region. An alternative name for it is unsupervised anomaly detection.
    
    \item \textbf{Novelty Detection: }Consider training data where there are no outliers, and we want to determine whether the incoming or fresh observation is an outlier or not. The outlier can be referred to as a novelty in this situation. Anomaly detection that is semi-supervised might be said to be involved.
    \item \textbf{Information Retrieval: } The purpose of this classification is to find samples in the unlabeled dataset that are related to the ones provided by the user.
    \item \textbf{One-vs-rest: } This is taken into account in this case when the negative class is too diversified and it is difficult to collect and label numerous negative samples \cite{sansone2018efficient}.
\end{itemize}
This technique can find outliers that deviate from the training set in some way. It is highly helpful in resolving a classification problem where samples for one class are in great abundance and samples for the other classes are scarce \citep{kang2007differentiated}. 
The primary objective of OCC is to estimate the support of the data distribution, which is very helpful in unsupervised learning, particularly in high-dimensional feature spaces where doing density estimation is exceedingly challenging \citep{sansone2018efficient}. Several real problems, including novelty discovery, outlier detection, and others.
Among the first authors to develop OCC algorithms were \cite{scholkopf1999support} and \cite{tax2004support}. A classifier that finds the least radius hypersphere enclosing data is proposed by \cite{tax2004support} whereas \cite{scholkopf1999support} specifically suggests a classifier that finds the hyperplane separating data from the origin with the largest margin. \cite{scholkopf1999support} establishes that despite these two approaches having differences between them, they both yield the same results for translation-invariant kernels like the Gaussian kernel. 
With the aim of addressing large-scale non-linear optimization problems, first-order approaches for non-linear optimization problem solving have been developed and used. Several methods that require the solution of linear systems of equations, such as the Exterior-Point Method (EPM) and Interior-Point Methods (IPM) \citep{griva20081, vanderbei1999interior} have been used to address quadratic programming issues. One of these techniques, for instance, works well for medium-sized situations with a few thousand variables

In recent years, algorithms like machine learning and others like as \citep{griva2018convergence} have been used to solve nonlinear optimization problems involving very large scale variables or datasets.
\citep{bloom2016fast} suggested combining the Fast Projected Gradient Approach (FPGM) with the Argumented Lagrangian (AL) method to train SVMs that simulate large-scale convex quadratic optimization problems with linear constraints and straightforward bounds. The study omitted the AL-convergence FPGM's analysis, despite the encouraging results of the model. On the other hand, the convergence of the AL-FPGM was theoretically investigated in paper \citep{griva2018convergence}.

The three main contributions of this paper are (i) applying a technique based on fast projected gradient \citep{bloom2016fast} for training dual soft margin OCC-SVMs, (ii) creating and implementing an AL-FPGM-based QP solver in Python for training OCC-SVMs, and (iii) testing the QP solver by training OCC-SVMs on few datasets used under PU-Learning problems \citep{sansone2018efficient}

The remaining parts of this paper is organized as follows: Section \ref{sec2} describes the dual soft margin OCC-SVMs training problem, Section \ref{sec3} describes the augmented Lagrangian method,
Section \ref{sec4} describes the fast projected gradient method, Section \ref{sec5} presents numerical results for training the  OCC-SVMs with the AL-FPGM and Section \ref{sec6} presents concluding remarks.

\section{The dual soft-margin SVM problem}\label{sec2}
\cite{scholkopf1999support} developed a method called "one-class classification" that extends the SVM methodology to handle training with just positive input.
Only positive data can be used with the suggested Schölkopf mechanism. The algorithm checks for "outliers" within the positive instances and uses them as negative examples \citep{manevitz2001one}. 
After changing the feature via a kernel, the one-class classification method of \cite{scholkopf1999support} treats the origin as the unique member of the second class. The image of one class is then separated from the origin using "relaxation parameters." Following that, the conventional OCC-SVMs algorithms are used \cite{manevitz2001one}.\\
The following quadratic programming problem must be solved in order to separate the data set from the origin:
	\begin{equation}
		\min \frac{1}{2}\|w\|^2 + \frac{1}{vn}\sum_{i=1}^{n} \zeta_i -\beta \label{svm_occ1}
	\end{equation}
	subject to:
	\begin{equation*}
		(w\cdot\phi(x_i))\geq \beta - \zeta_i \hspace{0.3cm} i=1,2,\dots,n \hspace{0.2cm} \zeta_i\geq 0
	\end{equation*}
Here, $v\in(0,1)$ is a parameter whose meaning will become clear later. Since nonzero slack variables $\zeta_i$ are penalized in the objective function, we can expect that if w and $\beta$ solve this problem, then the decision function $f(x)= sign(( w\cdot\phi(x))-\beta)$ will be positive for most examples $x_i$ contained in the training set, while the SV type regularization term $\|w\|$ will still be small. The actual trade-off between these two goals is controlled by $v$. It is possible to demonstrate that the solution has an SV expansion by deriving the dual problem and applying the kernel transformation.

Using Lagrange multiplier, constraint optimization problem can further be expressed as Equation (\ref{svm_occ2}).
\begin{align}
\begin{split}
    \mathcal{L}(w, \zeta, \alpha, b) &=\frac{1}{2}\|w\|^2 +  \frac{1}{vn}\sum_{i=1}^{n} \zeta_i -b - \sum_{i=1}^{n} \alpha_i((w\cdot\phi(x_i)-b)+ \zeta_i) \label{svm_occ2}
\end{split}
\end{align}
The Lagrange multiplier $\alpha$ must be greater or equal to zero. For the purpose of simplification, we expressed the $\frac{1}{vn}$ as $C$.
We further find the partial derivatives of the loss function $\mathcal{L}$ with respect to $\zeta, w \text{ and } \beta$
\begin{align}
  \frac{\partial \mathcal{L}}{\partial w} &= w- 
  \sum_{i=1}^{n}\alpha_ix_i=0 \implies w =\sum_{i=1}^{n}\alpha_ix_i\\
  \frac{\partial \mathcal{L}}{\partial \beta} &= -1 + \sum_{i=1}^{n}\alpha_i=0 \implies \sum_{i=1}^{n}\alpha_i = 1 \\
	 \frac{\partial \mathcal{L}}{\partial \zeta} &= C - \alpha_i=0 \implies C = \alpha_i
\end{align}
The optimal value will be obtained using $\zeta, w \text{ and } \beta$ from the equations above, and this gives rise to Equation \ref{svm_occ3}.
\begin{align}
 \mathcal{L}(w,\beta,\zeta,\alpha) &= \frac{1}{2}w^Tw + C\sum_{i=1}^{n}\zeta_i -\beta -w\sum_{i=1}^{n}\alpha_ix_i + \beta \sum_{i=1} \alpha_i- \sum_{i=1}^{n}\alpha_i\zeta_i \label{svm_occ3}\\
&=\frac{1}{2}w^Tw -w^Tw\\
\max \mathcal{L}(\alpha) &= -\frac{1}{2}\sum_{i=1}^{n}\sum_{j=1}^{n}\alpha_i\alpha_jx_i^Tx_j \label{eqn:occ_lmax}
\end{align}
Where $x^Tx = K(x,x)$ is a kernel to be computed.
Minimizing Equation (\ref{eqn:occ_lmax}) gives:
\begin{equation}
    f(\alpha) = \frac{1}{2}\sum_{i=1}^{n}\sum_{j=1}^{n}\alpha_i\alpha_jx_i^Tx_j \label{eqn:occ_lmin}
\end{equation}
Equation \ref{eqn:occ_lmin} can further be expressed in a more compact form as:
\begin{equation}
    f(\alpha) = \frac{1}{2}\alpha^TK(x,x)\alpha \label{eqn:comapct_lmin}
\end{equation}
The patterns $x_i$ with nonzero $\alpha_i$ are called SVs, where the coefficients are found as the solution of the dual problem:
    \begin{equation}
    f(\alpha) = \frac{1}{2}\alpha^TK(x,x)\alpha \hspace{0.2cm} \text{subject to} \hspace{0.2cm} \hspace{0.2cm} \sum_{i}^{n}\alpha_i=1 \label{svm_occfunc}
    \end{equation}
and the bounded set: 
\begin{equation*}
    Box = \{\alpha \in \mathbf{R}^m:  0\leq \alpha_i\leq \frac{1}{vn}, \hspace{0.5cm}i=1,\cdots,m\}
\end{equation*}
Then the optimization problem \ref{svm_occfunc} can be rewritten as follows:
\begin{equation}
    \min f(\alpha) \hspace{0.2cm} \text{s.t} \hspace{0.2cm} h(\alpha)=0 \label{svm_compact}
    \end{equation}

The augmented Lagrangian can be written as  follows:
\begin{equation}
    \mathcal{L}(\alpha) =f(\alpha) + \mu h(\alpha) + 0.5ch(\alpha)^2 \label{eqn:aug_lag}
\end{equation}
where $\mu\in\mathbb{R}$ is the unknown Langrage multiplier that corresponds to the equality constraint and $c>0$ is the scaling parameter.

\section{Augmented Lagrangian Method Algorithm}\label{sec3}
the augmented Lagrangian method constitute up of a sequences of approximate minimizations of $\mathcal{L}_c(\alpha, \mu)$ in $\alpha$  on the Box set
\begin{equation}
    \hat{\alpha} \approx \alpha(\hat{\mu}) = \arg\min_{\alpha\in Box}\mathcal{L}_c(\alpha, \mu) \label{eqn:al_equation}
\end{equation}
followed by updating the Lagrange multiplier $\mu$.
We use the following function, which measures the first-order optimality conditions for problem (\ref{eqn:al_equation}), as the stopping criteria:

Algorithm \ref{alg:method} provides a preliminary sketch of an augmented Lagrangian technique.
\begin{algorithm}
\caption{Augmented Lagrangian Methods}\label{alg:method}
\begin{algorithmic}
\State \textbf{Step 1:  Set} $\alpha^*:=\alpha=0, \text{ } \mu:=0, \text{ } rec:= acc(\alpha,\mu)$\\
\hspace{2cm}\textbf{Select} $c>0, \text{ }, 0<\theta<1, \text{ } \delta >1$

\State \textbf{Step 2: Find} $\hat{\alpha} \approx \text{arg}\min_{\alpha\in Box}\mathcal{L}(\alpha, \mu)$\\
\hspace{2cm} such that $\mu(\alpha,\mu) \leq \theta * rec$ using FPGM
\State \textbf{Step 3: Find} $\hat{\mu} \approx \mu + ch(\alpha)$
\State \textbf{Step 4: Set} $\alpha:=\hat{\alpha}, \epsilon:=\theta\epsilon, c:=\delta c$

\State \textbf{Step 5: If} $rec > \varepsilon$  then Goto 2. 

\State \textbf{Step 6: End}
\end{algorithmic}
\end{algorithm}

\noindent Standard QP methods can be used to solve this problem \ref{svm_occfunc}. The offset $\beta$ may be obtained by exploiting that for any $\alpha_i$ which is not at the upper or lower bound, the corresponding pattern $x_i$ satisfies:
\begin{equation}
    \beta=(w\cdot x_i)=\sum_{j}\alpha_jk(x_j,x_i)
\end{equation}
Note that the upper limits of the Lagrange multipliers go to infinity if $v\mapsto 0$.

\section{Fast Projected Gradient Method (FPGM)}\label{sec4}
The Lipschitz constant $L>0$ of the gradient of $\mathcal{L}_c$ must be estimated for the fast projected gradient method (FPGM) for the inequality 
\begin{equation}
    \|\nabla_\alpha\mathcal{L}_c(\alpha_{_1},\lambda)-\nabla_\alpha\mathcal{L}_c(\alpha_{_2},\lambda)\| \leq L\|\alpha_{_1}-\alpha_{_2}\|
\end{equation}
to hold $\forall \alpha_1, \alpha_2 \in \mathbb{R}^n$

 \vspace{0.2cm}
\noindent The gradient and the Hessian of $\mathcal{L}_c(\alpha,\lambda)$ for the OCC-SVM can be written as follows:
\begin{align}
\begin{split}\label{grad-pufpgm}
    \nabla_\alpha\mathcal{L}_c(\alpha,\lambda) &= K\alpha - (\lambda  -c(\alpha^Te-1))e^T
\end{split}
\end{align}

\begin{equation}
     \nabla_{\alpha\alpha}^2\mathcal{L}_c(\alpha,\lambda) =  K  +cee^T
\end{equation}
    where K is an $n \times n$ matrix,
\begin{equation*}
    \begin{split}
        K_{i,j} &=k(x_i,x_j)\\
        e &= (1,...,1)^T\in \mathbb{R}^{n}
    \end{split}
\end{equation*}
$\mathcal{L}_c$ is a quadratic form with respect to $\alpha$
\begin{equation}
    L= \|\nabla^{2}_{\alpha\alpha} \mathcal{L}_c(\alpha, \lambda)\|= \|K + cee^T\|
\end{equation}
where the biggest singular value of a matrix is the matrix spectral norm, i.e the constant that is dependent only on the kernel Matrix \textbf{K} and the scaling parameter c \citep{bloom2016fast}.\\
In the terms of Gaussian kernel we have this expression:
\begin{align}
    \mathcal{L}\leq trace (\nabla^2_{\alpha\alpha}\mathcal{L}_c (\alpha,\lambda) &= trace(K + cee^T)\\
    &= trace(K)+trace(cee^T)
\end{align}
In the estimation above, we took into account the facts that the trace of a matrix is the sum of its eigenvalues and that, when all eigenvalues are nonnegative, the biggest eigenvalue is either less than or equal to the total of all eigenvalues.\\
Therefore, the estimation ($L \approx trace(K)+trace(c(ee^T)$) is considered in the training.
Similar kernel-specific bounds for additional kernels can be determined. For the dot product kernel, polynomial kernel and others, for example, the same estimation ($L \approx trace(K)+trace(cee^T)$) can be applied.
The most computational expensive term of the $\mathcal{L}_c(\alpha, \lambda)$ is the matrix-vector product $K\alpha$ calculation which carries $\mathcal{O}(n^2)$ basic operations in the case when the matrix under consideration is dense (Steps 3 and 7 in Algorithm \ref{alorith:fpgm}).
The projection operator $P_{Box} : \mathbb{R}^n \mapsto  Box$ (Step 3) is not computationally expensive (see algorithm \ref{pbox:method}) and expects only $\mathcal{O}(n)$ basic arithmetic operations. Less than dozen of arithmetic operations are contained in the remaining stages together. Therefore, one iteration of FPGM expects $\mathcal{O}(n^2)$ operations.
Algorithm \ref{alorith:fpgm} describes the FPGM used in the step 2 of the Algorithm \ref{alg:method}. 
For the sequences of $\{\alpha_s\}$ that are generated by the FPGM for a convex quadratic problem, the following bound hold \citep{bloom2016fast}
\begin{equation}
    \mathcal{L}_c(\alpha_s, \lambda) -  \mathcal{L}_c(\alpha(\lambda), \lambda) \leq
    \frac{2L\|\alpha_0 - \alpha(\lambda) \|^2}{(s+1)^2}
\end{equation}
where $L>0$ is the Lipschitz constant and $\alpha(\lambda)$ is the expected solution of the problem (\ref{eqn:al_equation}) under consideration.  As a result, the FPGM converges to the  minimum value augmented Lagrangian with an error order of $\mathcal{O}(s^{-2})$, where s is the number of iterations. Furthermore, with reference to the estimation of L for the various kernels and the fact that both $\alpha_0$ and $\alpha(\lambda) \in Box$, the  number of steps expected for the FPGM to converge to the minimum of the Augmented Lagrangian $\mathcal{L}_p(\alpha(\lambda),\lambda)$ can be estimated \citep{bloom2016fast}.\\
Since $\{\mathcal{L}(\alpha)\}$ is a decreasing sequence (with reference to the fact that $\alpha^k \not= \alpha^{k+1} \forall k \geq 0$ and that the objective function is minimized at every iteration) and also bounded below (based on the existence of an unknown global optimum), it can be said that convergences. Cauchy sequence can also be applied to prove that it is convergent.\\
 By using this fact and by considering that $\| \mathcal{L}(\alpha^k) - \mathcal{L}(\alpha^{k+l})\| > \|\alpha^k - \alpha^{k+l} \|_2 \forall k,l \geq 0$, It can be inferred that $\{\alpha^k\}$
Therefore since the sequence lies also in a closed feasible set, it is convergent. That's as $k\mapsto \infty$, $\alpha^k \mapsto \alpha^{k+1}$ then, the algorithm (1) yields a sequence of points.

\begin{algorithm}
\caption{Operator $P_{Box}$: Projection of $\alpha\in\mathbb{R}^n$ onto the set B}\label{pbox:method}
\begin{algorithmic}
\State \textbf{1: }Loop over all $i=1,\dots,n$
\State \textbf{2: If} $\alpha_i < 0$ then Set $\alpha_i =0$
\State \textbf{3: If} $\alpha_i > C$ then Set $\alpha_i =C$
\State \textbf{4:} Return $\alpha_1$
\end{algorithmic}
\end{algorithm}


\begin{algorithm}
\caption{Fast Projected Gradient Method}\label{alorith:fpgm}
\begin{algorithmic}
\State \textbf{1:  Input} $(\alpha,\lambda)$
\State \textbf{2:  Set} $\Bar{\alpha} = \alpha, t=1. \text{ Select } L>0.$
\State \textbf{3: Set} $\hat{\alpha} = P_{Box}(\alpha-\frac{1}{L}\nabla_\alpha\mathcal{L}_{k}(\alpha,\lambda)$
\State \textbf{4: Set} $\Bar{t}=0.5(1+\sqrt{1+4t^2}$
\State \textbf{5: Set} $\alpha=\hat{\alpha}+\frac{t-1}{\Bar{t}}(\hat{\alpha}-\Bar{\alpha})$
\State \textbf{6: Set} $\Bar{\alpha}:=\hat{\alpha}, t:=\Bar{t}$
\State \textbf{7: If} $rec > RequiredAccuracy$ then Goto 2. 
\State \textbf{8: Output} $\hat{\alpha}$
\end{algorithmic}
\end{algorithm}
The AL method falls into the class of the proximal algorithms that have been extensively investigated in the papers of \cite{rockafeller1976monotone,rockafeller1976augmented}. One of possible AL-FPGM convergence analysis paths is to use of the theory developed in \cite{rockafeller1976monotone}. The importance of using the proximal-point theory is the possibility to demonstrate the convergence of the AL-FPGM for any $c>0$; which would explain the possibility of obtaining convergence for small $c$.

\section{Numerical results and discussion}\label{sec5}
In this section, we report on the numerical results of training OCC-SVMs with the AL-FPGM on datasets that were used to solve PU-Learning problem \citep{sansone2018efficient}. To assess an improvement in efficiency of the AL-FPGM relatively to optimization methods that solve linear systems at every iteration, we compared the efficiency of the former to the OCC-SVM solver in python. It was observed that both approaches converges in less than 1 second. We selected the OCC-SVM solver in Python for a few reasons. First, it happens to be the widely used solver built for OCC-SVM. Secondly, it turns out to be an efficient and accurate method that solves linear systems at each step while keeping the number of iterations low. Lastly, the numerical results of testing the OCC-SVM solver is published and available for comparison. The OCC-SVM implemented in Python was developed for anomaly detection. For the training set, $25\%$ of the positives were used whiles the remaining $75\%$ of the positives and all the negatives were used for the testing. 
The Gaussian kernel $K(x_i,x_j) = e^{-\gamma\|x_i-x_j\|}$ with $\gamma$ values of 0.1, 0.5 and 1 were used and the results were presented in Table \ref{tab:python_alfpgm}. For the AL-FPGM we used the scaling parameter value $c=0.1$  and the stopping criteria of $10^{-6}$. The other parameters used were $\theta = 0.99$ and $\delta = 1.01$ for all the runs.
\centering
\begin{table}
\caption{Results obtained in using the two method under One Class Classification-SVM}
   \begin{adjustbox}{width=11cm, center}
    \begin{tabular}{l |c c c ||c c c }\hline
    & \multicolumn{3}{c}{Python } &  \multicolumn{3}{c}{AL-FPGM }\\\cline{2-4}\cline{5-7}
    Dataset & \multicolumn{3}{c}{$\gamma - values$}&
    \multicolumn{3}{c}{$\gamma - values$} \\\cline{2-4}\cline{5-7}
			& 0.1 & 0.5 & 1.0 & 0.1 & 0.5 & 1.0 \\\hline
    Australian & 61.0 & 59.8 & 58.7 & 66.7 & 66.7 & 66.7 \\\hline
    Clean1 & 53.1 & 42.4 & 43.3 & 49.6 & 49.6 & 49.6 \\\hline
    Diabetes & 72.9 & 73.1 & 73.2 & 71.4 & 71.4 & 71.4 \\\hline
    Heart & 66.7 & 68.0 & 69.5 & 85.1 & 85.1 & 85.1 \\\hline
    Heart-Statlog & 70.6 & 71.8 & 72.3 & 80.0 & 80.0 & 80.0 \\\hline
    House & 86.6 & 69.7 & 63.9 & 91.1 & 91.1 & 91.1 \\\hline
    House-Votes & 70.3 & 77.4 & 78.4 & 80.1 & 80.1 & 80.1 \\\hline
    Ionosphere & 47.1 & 46.3 & 46.3 & 58.5 & 58.5 & 58.5 \\\hline
    Isolet & 92.3 & 59.8 & 36.4 & 93.8 & 93.8 & 93.8 \\\hline
    Krvskp & 46.5 & 43.5 & 43.9 & 43.5 & 43.5 & 43.5 \\\hline
    Liver Disorders & 45.4 & 45.2 & 43.1 & 41.6 & 41.6 & 41.6 \\\hline
    Spectf & 46.2 & 45.8 & 47.2 & 49.7 & 49.7 & 49.7 \\\hline
    \end{tabular}
     \end{adjustbox}
    \label{tab:python_alfpgm}
\end{table}

\section{Conclusion}\label{sec6}
\noindent 
This manuscript presents numerical results on training dual soft margin OCC-SVMs with AL-FPGM. We developed and implemented a QP solver based on the AL-FPGM and tested the QP solver by training medium-sized data sets from the USMO paper. The numerical results presented here indicate that the AL-FPGM is an efficient training algorithm for medium-sized SVMs \cite{bloom2016fast} and OCC-SVMs. The numerical results demonstrate that starting with a sample size of about a few  data points, AL-FPGM consistently outperforms the original OCC-SVMs solver in Python in almost all the datasets. Therefore, we believe that the AL-FPGM has a good potential to help in improving classifications results. With reference to the dataset used and the performance, it can be inferred that instead of dealing with larger matrices in PU-learnig problems OCC-SVMs with AL-FPGM. A significant flaw in the AL-FPGM is the obligation to work with all the variables at once. The algorithm's current state makes it difficult for it to compete with other similar methods. It is important to note that the presented AL-FPGM can still be used as a starting point for the development of a more complex algorithm for training OCC-SVMs with a large amount of data (hundreds of thousands of data points), despite the fact that it does not use any type of sequential incorporation of the training data. As a result, we think the AL-FPGM has a strong potential of complementing other current techniques for training large OCC-SVMs. We intend to address the problem of storing large Gram matrices under OCC-SVMs in the future. The algorithm's memory requirements may be reduced when dealing with large Gram matrices. The numerical findings obtained support the theory and demonstrate that the AL-FPGM technique in the OCC-SVM frameworks has a strong chance of succeeding as a competitive tool in the NLP area.

\bibliography{sn-article}


\end{document}